\documentclass[10pt,twocolumn,letterpaper]{article}

\usepackage{iccv}
\usepackage{times}
\usepackage{epsfig}
\usepackage{graphicx}
\usepackage{amsmath}
\usepackage{amssymb}
\usepackage{authblk}

\usepackage[breaklinks=true,bookmarks=false]{hyperref}

\iccvfinalcopy 

\def\iccvPaperID{7} 

\ificcvfinal\fi

\begin{document}

\title{Fashion Image Retrieval with Capsule Networks}
\author{Furkan K{\i}nl{\i}\textsuperscript{1}}
\author{Bar{\i}\c{s} \"{O}zcan\textsuperscript{2}}
\author{Furkan K{\i}ra\c{c}\textsuperscript{3}}

\affil{\"{O}zye\u{g}in University, Turkey \\
{\tt\small \{\textsuperscript{1}furkan.kinli, \textsuperscript{3}furkan.kirac\}@ozyegin.edu.tr ,} {\tt\small \textsuperscript{2}baris.ozcan.10097@ozu.edu.tr}
}

\maketitle
\thispagestyle{empty}

\begin{abstract}
   In this study, we investigate in-shop clothing retrieval performance of densely-connected Capsule Networks with dynamic routing. To achieve this, we propose Triplet-based design of Capsule Network architecture with two different feature extraction methods. In our design, Stacked-convolutional (SC) and Residual-connected (RC) blocks are used to form the input of capsule layers. Experimental results show that both of our designs outperform all variants of the baseline study, namely FashionNet, without relying on the landmark information. Moreover, when compared to the SOTA architectures on clothing retrieval, our proposed Triplet Capsule Networks achieve comparable recall rates only with half of parameters used in the SOTA architectures.
\end{abstract}

\section{Introduction}

Fashion has recently become one of the most featured topics of interdisciplinary studies in Computer Science. With the emergence of deep learning based solutions, fashion-related researches start to get promising results on various subjects including clothing recognition, attribute prediction, clothing retrieval, body segmentation, and style prediction. Retrieving the desired clothing image from a collection is one of the most challenging tasks in fashion domain, and it is attacked by such a mechanism that learns to capture different notions of the similarities between the images in a common subspace. 

There has been numerous studies \cite{wtbi, darn, deepfashion, weakly-anno, vam, hdc, bier, htl, abe} to employ Convolutional Neural Networks (CNNs) to their solutions. However, CNNs, by their nature, have some limitations such as losing the hierarchical spatial information of the objects and not being robust to affine transformations. Recently, an alternative deep learning architecture, namely \textit{Capsule Networks}, and a novel \textit{dynamic} routing algorithm have been proposed by Sabour and Hinton \etal \cite{capsule}. In this design, with the help of the routing-by-agreement algorithm, it is possible to learn more descriptive information about the objects without losing the intrinsic spatial relationship between the object and its parts. Therefore, Capsule Networks have the capacity for recognizing the images regardless of the visual angle and without requiring different transformations, since this architecture can inherently learn higher dimensional pose configuration of the images. 

\begin{figure}[t]
\begin{center}
\includegraphics[width=1\linewidth]{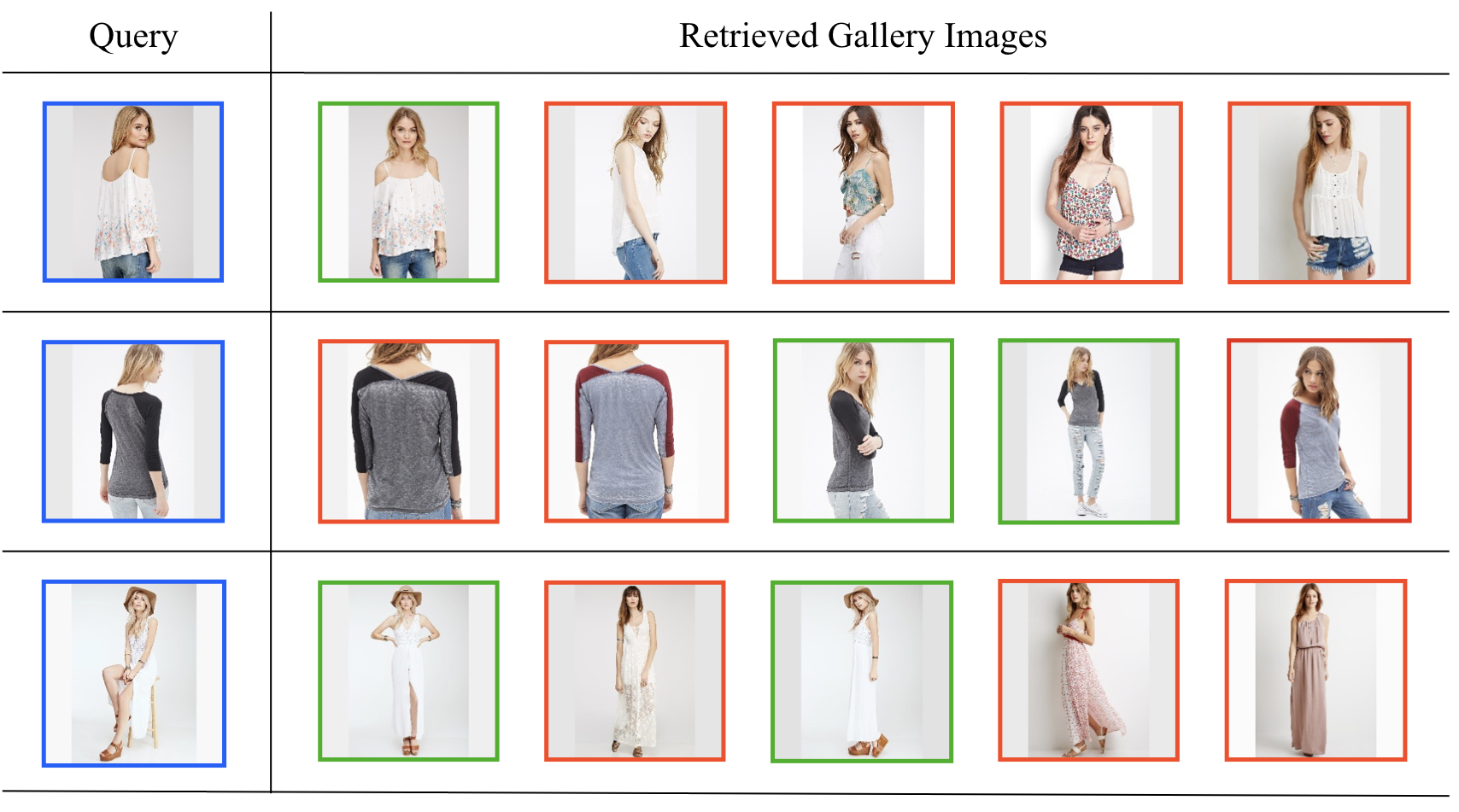}
\end{center}
\caption{Some examples of retrieved images by our architectures. Blue: query, Green: correct, Red: wrong.}
\label{fig:datasetQuery}
\end{figure}

In this study, we employ Capsule Networks to clothing retrieval problem by extending their capabilities with some improvements. First, we extract the features of larger-sized clothing images by more powerful methods (\ie stacked or residual-connected convolutional layers), and forward these features to fully-connected capsules. Next, we introduce a Triplet-based design of Capsule Networks that learns the similarity between triplets. Lastly, we train our proposed architectures on in-shop partition of DeepFashion data set \cite{deepfashion}, and compare our results with the baseline study, namely \textit{FashionNet} \cite{deepfashion} and the other SOTA methods.

\section{Related Works}

\begin{figure*}[ht!]
\begin{center}
\includegraphics[width=0.9\linewidth]{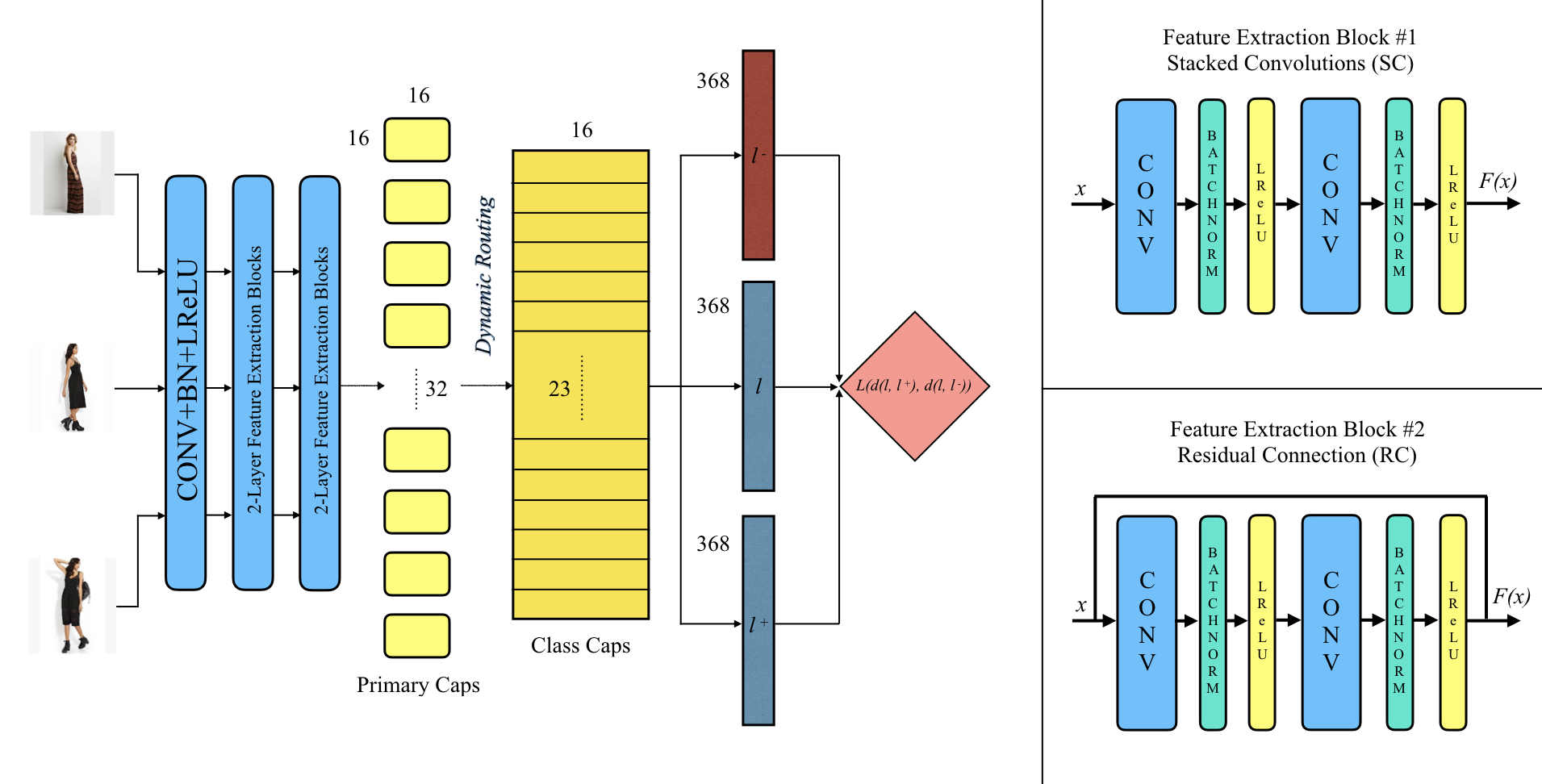}
\end{center}
\caption{Illustration of our proposed architectures containing different feature extraction blocks.}
\label{fig:architecture}
\end{figure*}

Clothing retrieval has become more important after some major developments in Computer Science and the emergence of e-commerce. Recent studies \textit{generally} attack to this task by using deep convolutional networks. \cite{wtbi} introduces an excessively challenging task, namely \textit{Exact  Street  to  Shop}, where the goal is to match the exact same item in the photos captured by users to online shopping photos. \cite{darn} proposes Dual Attribute-aware Network (DARN) to address the cross-domain image matching problem. \cite{deepfashion} introduces a new data set, namely \textit{DeepFashion}, which has a vast  amount  of  large-scale  clothing  images  annotated  with  numerous  attributes, landmark  information  and  cross-domain  image  correspondences. \cite{weakly-anno} demonstrates that integrating bag-of-words approach to weakly-supervised learning process can achieve promising results on clothing retrieval task. \cite{vam} proposes a Visual Attention Model (VAM), and introduces a novel Dropout-like connection after attention layers. \cite{hdc} addresses the issues of defining a model with right complexity and choosing hard samples carefully during training.  \cite{bier}  shows  how  to  improve  the  robustness  of  the  feature embeddings  by  exploiting  the  independence  within  ensembles. \cite{htl}  introduces  hierarchical  triplet  loss  (HTL)  to  address  the random  sampling  issue  during  training  a  triplet  loss. \cite{abe}  proposes  multiple-way  attention-based  ensemble  architecture that  learns  the  feature  embeddings  with  multiple  attention  masks.

\section{Methodology}

\subsection{Capsules}

Capsules are groups of neurons that convey higher dimensional information throughout the network in more refined way. This information is interpreted as the pose configuration and the existence probability of an instance. Each capsule in a higher level is formed by the routing of incoming votes from the capsules in lower level. At this point, these votes are calculated by the linear transformation of the pose configuration. During dynamic routing \cite{capsule}, the linear combination of incoming votes weighted by their coefficients (\ie coupling coefficients) forms the non-activated outputs in higher level capsules. For each iteration, the weights of these votes are updated with respect to the dot product of the incoming votes and the outputs in higher level capsules. This is called \textit{agreement between capsules}. Finally, the output of each capsule in lower level is determined by \textit{squashing} function as proposed in \cite{capsule}.

\subsection{Proposed Architectures}

In our design, we adjust the original Capsule Network structure to a Triplet-based version, so that the network can learn the similarity between two images by feeding the objective function with the embedded representations extracted by capsules. At this point, our Capsule Network design aims to minimize the Triplet loss shown in Equation \ref{eq:loss}, where $d$ is the Euclidean distance metric, $\alpha$ is distance margin, $l$, $l^{+}$, $l^{-}$ are the latent capsule embeddings extracted from the anchor image $x$, positive image $x^{+}$ and negative image $x^{-}$ respectively. During forming these embeddings, we normalize latent capsules by L2-norm, and then we mask all capsules but the one that belongs to the correct class to zero.

\begin{equation}
\sum_{i}  [d(l_{i}, l_{i}^{+}) - d(l_{i}, l_{i}^{-}) + \alpha]
\label{eq:loss}
\end{equation}

As illustrated in Figure \ref{fig:architecture}, Capsule Networks essentially contain two main blocks: feature extraction block and capsule layers. There is only one feature extraction block that has a single convolutional layer with 64 filters in the original design proposed by Sabour and Hinton \etal \cite{capsule}. Extracting the features by such a shallow structure may be enough for one-channel handwritten digit images with the size of $28 \times 28$ \cite{capsule}. However, fully-connected capsules need more complex features to achieve better results on more complicated image-related problems. Therefore, we design two different feature extraction blocks to form more powerful features as the input of capsules. First, a number of convolutional layers are stacked without using any pooling operation between them, and the latter is to connect these layers as residual. In both of our designs, leaky form of linear rectifier \cite{leaky-relu} is used as activation function, and batch normalization \cite{bn} is applied between convolutional layers.

Furthermore, capsule layers are kept identical in both designs. There are two fully-connected capsule layers, namely \textit{Primary Capsule} and \textit{Class Capsule}. Primary Capsule is the layer where the extracted features are grouped with respect to the capsule dimensionality. In our designs, this layer has 32 channels of 16-dimensional capsules that are fully-connected to Class Capsule. Next, there are $c$ number of 16-dimensional capsules in Class Capsule layer, where $c$ is the number of classes in the data set. Activations and the latent capsule vectors of Class Capsule are calculated via dynamic routing with 3 iterations. Any kind of reconstruction methods (\eg as in \cite{capsule}) is not applied to our Capsule Network designs.

\section{Experiments}

The experiments for proposed Stacked-convolutional (SCCapsNet) and Residual-connected (RCCapsNet) architectures are conducted on in-shop partition of DeepFashion data set \cite{deepfashion}. Both are trained on 25k training images, and tests are performed by using 14k query and 12k gallery images. Since this task is an information retrieval task, the performance is measured by Recall@K metric, where K is 1 or multiplies of 10 up to 50. Moreover, as mentioned in Schroff \etal \cite{facenet}, negative hard sampling strategy improves the convergence behavior of the model significantly. Based on this strategy, the negative images are picked as the closest image to the anchor provided that they are of different categories; whereas we pick each possible positive image in the data set as the positive one.

As shown in Table \ref{tab:baseline}, SCCapsNet and RCCapsNet achieve better retrieval performance than all variants of the baseline study (\ie FashionNet) by a wide margin. It is important to note that both of our proposed architectures use only images during training in contrast to the baseline study where the network is supported by different number of attributes and the landmark information. These experiments demonstrate that our Capsule Network designs can inherently learn pose configuration of the objects without any requirement of recovering pose information.

\begin{table}[t]
    \caption{Recall@K performance of the variants of
    the baseline study \cite{deepfashion} and our proposed model.
    FashionNet has different building blocks where the model
    has different numbers of attributes (A) (i.e. 100, 500 and
    1000), or fashion landmarks (L) are replaced with human
    joints (J) or poselets (P). SCCapsNet and RCCapsNet do not use
    any extra side information during training.}
    \begin{center}
        \begin{tabular}{c|c|c}
             \hline \hline
            \textbf{Models} & \textbf{Top-20 (\%)} & \textbf{Top-50 (\%)}  \\
            \hline \hline
            FashionNet+100A+L   &  57.3 &    62.5   \\
            FashionNet+500A+L   &  64.6 &    69.5   \\ 
            FashionNet+1000A+J  &  68.0 &    73.5   \\ 
            FashionNet+1000A+P  &  70.0 &    75.0   \\ 
            FashionNet+1000A+L  &  76.4 &    80.0   \\ \hline 
            SCCapsNet \textit{(ours)}         &  81.8 &    90.9   \\ 
            RCCapsNet \textit{(ours)}         &  \textbf{84.6} &    \textbf{92.6}   \\ \hline
            \hline
        \end{tabular}
    \end{center}
    \label{tab:baseline}
\end{table}

Table \ref{tab:sota} summarizes in-shop clothing retrieval results of SCCapsNet, RCCapsNet, and the SOTA methods. These figures indicate how successful our proposed designs are, and what the main limitations of them are when compared to the SOTA CNN-based architectures. First, both of our designs outperform the earlier methods (\textit{i.e.} WTBI \cite{wtbi} and DARN \cite{darn}) which both disparately use semantic attributes to improve the overall performance, but neglect pose configurations of the images during training. According to Top-20 Recall@K scores, while SCCapsNet improves the scores of the best FashionNet variant by 31\% and 14\%, RCCapsNet has even better performance with a margin of 34\% and 17\% respectively. The other approach whose performance falls behind in ours is the method of leveraging weakly-annotated textual descriptors of the images proposed by Corbi\'ere \etal \cite{weakly-anno}. In this design, these textual descriptors (\textit{i.e.} bag-of-words) represent different coarse semantic concepts such as texture information, color and shape. Capsules can directly learn these concepts from the images in a sophisticated way, and hence, SCCapsNet and RCCapsNet can achieve higher Recall@K scores than this approach without taking advantage of bag-of-words descriptors.

\begin{table*}[ht!]
\caption{Experimental results of in-shop clothing retrieval task on DeepFashion data set. "-":  not reported.}
\begin{center}
\begin{tabular}{c|c|c|c|c|c|c|c}
 \hline \hline
\textbf{Models} & \textbf{\# of} & \textbf{Top-1} & \textbf{Top-10} & \textbf{Top-20} & \textbf{Top-30} & \textbf{Top-40} & \textbf{Top-50} \cr  & \textbf{Params (M)} &  \textbf{(\%)} &  \textbf{(\%)} &  \textbf{(\%)} & \textbf{(\%)} &  \textbf{(\%)} &  \textbf{(\%)} \\
\hline \hline
WTBI \cite{wtbi} & 60  & 35.0 & 47.0 & 50.6 & 51.5 & 53.0 & 54.5 \\ 
DARN \cite{darn} & 105 & 38.0 & 56.0 & 67.5 & 70.0 & 72.0 & 72.5 \\  
FashionNet \cite{deepfashion} & 134 & 53.2 & 72.5 & 76.4 & 77.0 & 79.0 & 80.0 \\
Corbi\'ere \textit{et al.} \cite{weakly-anno} & 25 & 39.0 & 71.8 & 78.1 & 81.6 & 83.8 & 85.6 \\ \hline 
SCCapsNet \textit{(ours)} & 2.5 & 32.1 & 72.4 & 81.8 & 86.3 & 89.2 & 90.9 \\ 
RCCapsNet \textit{(ours)} & 4.5 & 33.9 & 75.2 & 84.6 & 88.6 & 91.0 & 92.6 \\ \hline
HDC \cite{hdc} & 5 & 62.1 & 84.9 & 89.0 & 91.2 & 92.3 & 93.1 \\ 
VAM \cite{vam} & 6 & 66.6 & 88.7 & 92.3 & - & - & - \\ 
BIER \cite{bier} & 5 & 76.9 & 92.8 & 95.2 & 96.2 & 96.7 & 97.1 \\ 
HTL \cite{htl} & 5 & 80.9 & 94.3 & 95.8 & 97.2 & 97.4 & 97.8 \\
A-BIER \cite{bier} & 5 & 83.1 & 95.1 & 96.9 & 97.5 & 97.8 & 98.0 \\ 
ABE \cite{abe} & 10 & 87.3 & 96.7 & 97.9 & 98.2 & 98.5 & 98.7 \\ \hline
\hline
\end{tabular}
\end{center}
\label{tab:sota}
\end{table*}

In addition to all these, our proposed architectures cannot achieve the performances of more advanced CNN-based architectures. In these designs, there are various techniques applied to CNNs to boost the overall performance, which are alternative hard sampling strategies \cite{hdc}, more advanced objective functions \cite{htl,bier}, network ensembling \cite{bier, abe} and attention-based mechanisms \cite{vam,abe}. Although these techniques may significantly improve the overall performance in CNNs, in principle, they increase the model complexities by a wide margin, or increase training time considerably. At this point, the numbers of trainable parameters in SCCapsNet and RCCapsNet are respectively $\sim$2.5 and $\sim$4.5 million, while the SOTA methods have twice as many trainable parameters in their models. Capsule Networks need more time for training than CNNs since dynamic routing algorithm is a relatively slow routing mechanism when compared to the pooling variants. Therefore, within limited computational resources, these techniques are not yet applied to our models to boost the overall performance of our Capsule Network designs, and left as future research ideas.

\section{Conclusion}

In this study, we present two different Triplet-based designs of Capsule Networks with more powerful feature extraction blocks, and employ them to clothing retrieval task. Experiments show promising results where both of our designs outperform all FashionNet variants without any extra information besides to the images. Moreover, when compared to the SOTA methods, our designs perform comparably well with only the half of the number of parameters as in the SOTA methods, and it shows the potential of Capsule idea in case of the computational burdens are lightened.



{\small
\bibliographystyle{ieee}
\bibliography{egbib}
}

\end{document}